\documentclass{manuscript}
\usepackage{xcolor}

\usepackage{lipsum} 
\usepackage[margin=1in]{geometry}
\usepackage{multirow, array}
\usepackage{makecell}
\usepackage{amsmath}
\usepackage{tabularx}
\usepackage{wrapfig}

\begin{document}

\title{Beyond Feature Importance: Feature Interactions in Predicting Post-Stroke Rigidity with Graph Explainable AI}
\author{Jiawei Xu, MS$^{\S,1}$, Yonggeon Lee, MS$^{\S,1}$, Anthony Elkommos Youssef, MS$^2$, Eunjin~Yun,~MS$^1$, Tinglin Huang, MS$^3$, Tianjian Guo, PhD$^4$,\\ Hamidreza Saber, MD, MPH$^5$, Rex~Ying,~PhD$^3$, Ying Ding, PhD$^{*1,5}$}
\institutes{
    $^1$ School of Information, UT Austin, Texas;
    $^2$ College of Natural Sciences, UT Austin, Texas;
    $^3$ Department of Computer Science, Yale University, Connecticut;\\
    $^4$ McCombs School of Business, UT Austin, Texas;
    $^5$ Dell Medical School, UT Austin, Texas \\
    $^\S$ Co-first authors.
    $^*$Corresponding Author. Email: ying.ding@ischool.utexas.edu
}

\maketitle

\section*{Abstract}
\textit
{This study addresses the challenge of predicting post-stroke rigidity by emphasizing feature interactions through graph-based explainable AI. Post-stroke rigidity, characterized by increased muscle tone and stiffness, significantly affects survivors' mobility and quality of life. Despite its prevalence, early prediction remains limited, delaying intervention. We analyze 519,000 stroke hospitalization records from the Healthcare Cost and Utilization Project dataset, where 43\% of patients exhibited rigidity. We compare traditional approaches such as Logistic Regression, XGBoost, and Transformer with graph-based models like Graphormer and Graph Attention Network. These graph models inherently capture feature interactions and incorporate intrinsic or post-hoc explainability. Our results show that graph-based methods outperform others (AUROC 0.75), identifying key predictors such as NIH Stroke Scale and APR-DRG mortality risk scores. They also uncover interactions missed by conventional models. This research provides a novel application of graph-based XAI in stroke prognosis, with potential to guide early identification and personalized rehabilitation strategies.}

\begin{figure}[H]
    \centering
    \includegraphics[width=1\linewidth]{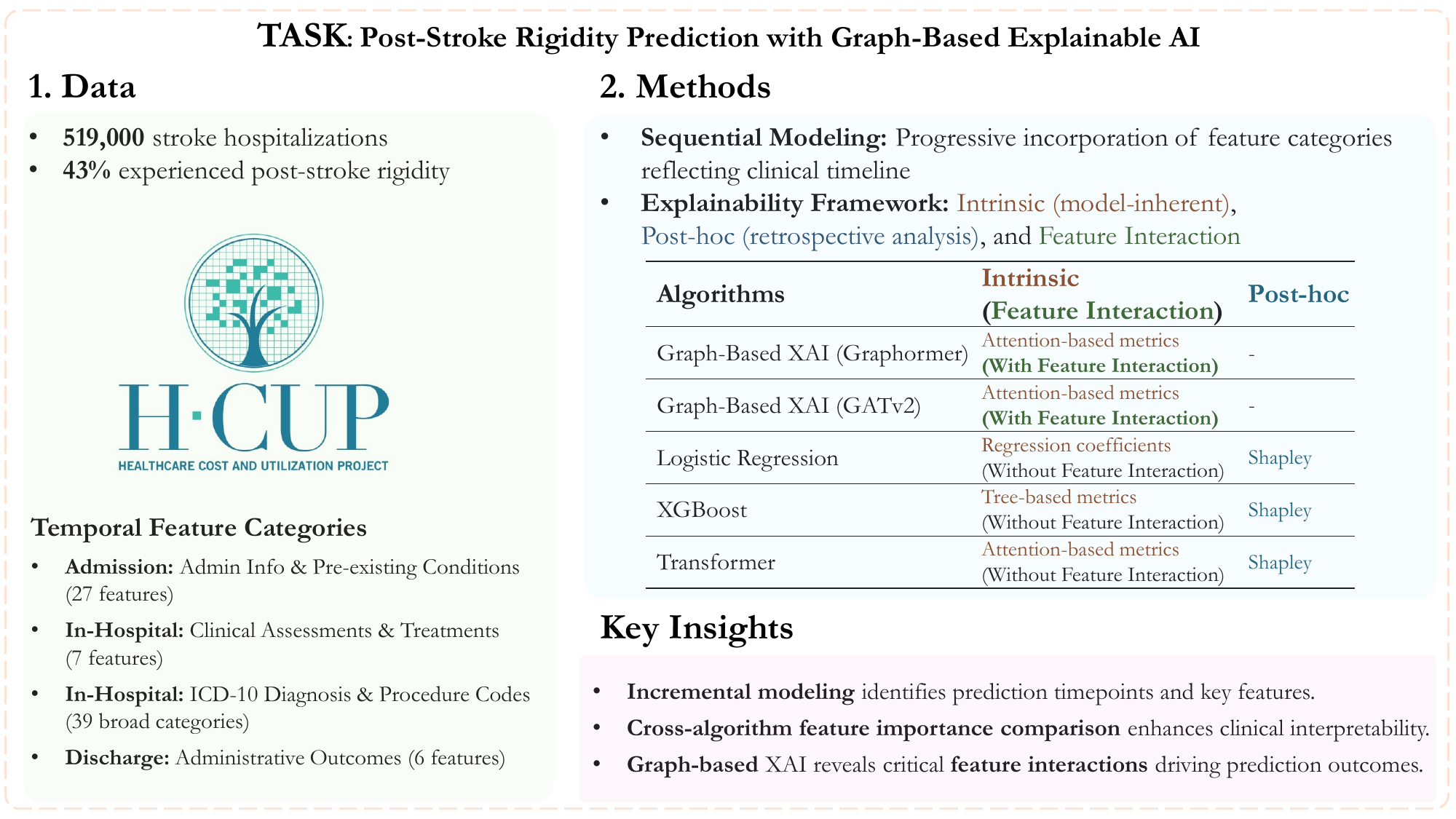}
    \caption{Diagram of the Beyond Feature Importance: Feature Interactions in Predicting Post-Stroke Rigidity with Graph Explainable AI.}
    \label{fig:research_design}
\end{figure}

\section*{Introduction}

Stroke is a leading cause of long-term disability worldwide, with over 12 million new cases annually~\cite{gorelick2019global}. Among stroke survivors, post-stroke rigidity, a condition characterized by increased muscle tone and stiffness, significantly impairing mobility~\cite{miyazaki2024logistic}, quality of life~\cite{martino2020health}, and rehabilitation outcomes~\cite{lin2018predicting}. Despite its prevalence, no widely used clinical tool effectively predicts which patients will develop post-stroke rigidity, leading to delayed interventions and suboptimal treatment outcomes. Early identification of at-risk patients is essential, as timely therapeutic strategies, such as neurotoxin injections and specialized physiotherapy, can mitigate rigidity severity and improve recovery trajectories~\cite{glaess2021early, bavikatte2021early}. However, rigidity prediction remains a challenging task due to the complex interplay of neurological, physiological, and demographic factors.
Existing research on post-stroke outcome prediction has explored various machine learning and deep learning approaches, including logistic regression, decision trees, and neural networks, often incorporating explainability techniques such as SHAP and LIME to enhance model interpretability~\cite{lee2023prediction, ji2023predicting, lv2023interpretable, lee2024machine}. However, these studies predominantly focus on feature importance at an individual level, often neglecting feature interactions, which can play a crucial role in clinical decision-making. Furthermore, most prior work is constrained by relatively small sample sizes or single-center datasets, limiting the generalizability of findings~\cite{ge2019predicting, li2024feature, cox2016predictive}. Additionally, while graph-based learning has shown promise in handling structured medical data for tasks such as heart failure prediction and chronic disease diagnosis~\cite{boll2024graph}, its application in post-stroke rigidity prediction remains largely unexplored.

To address these limitations, this study leverages a large-scale dataset of 519,000 stroke hospitalizations from the Healthcare Cost and Utilization Project (HCUP-US), where 43\% of hospitalizations exhibited rigidity. We develop a predictive framework that integrates multiple machine learning models, including traditional algorithms such as Logistic Regression and XGBoost, deep learning architectures like Transformers, and intrinsically interpretable graph-based models such as Graphormer and Graph Attention Network (GATv2)~\cite{guo2024explainable}. Our contributions are three-fold:

\begin{itemize}
    \item Incremental modeling for temporal insights: We identify critical time points for rigidity prediction, providing a structured understanding of disease progression.
    \item Cross-algorithm feature importance comparison: By evaluating feature contributions across different modeling approaches, we enhance clinical interpretability and robustness.
    \item Graph-based XAI for feature interaction analysis: Unlike traditional models that focus on individual feature importance, our approach explicitly captures feature interactions, offering deeper insights into the underlying factors driving post-stroke rigidity.
\end{itemize}

By systematically evaluating various explainability mechanisms and leveraging graph-based modeling techniques, this study advances the interpretability and predictive capabilities of post-stroke rigidity models. Our findings provide a novel perspective on the role of feature interactions in clinical prediction tasks, with potential implications for improving early intervention strategies and personalized stroke rehabilitation.

\section*{Related Work}
\paragraph{\textit{Post-Stroke Healthcare Outcome Prediction}\\}

Machine learning and deep learning, and explainable AI (XAI) techniques, have been applied to predict post-stroke rigidity and spasticity~\cite{glaess2021early,cox2016predictive}, as well as other outcomes such as ADL~\cite{lin2018predicting}, cognitive impairment~\cite{lee2023prediction,ji2023predicting}, gait independence~\cite{miyazaki2024logistic}, and mortality~\cite{wang2022risk}. Approaches include logistic regression, random forests, deep neural networks, and convolutional neural networks, with sample sizes ranging from as few as six~\cite{li2024feature} to over 10,000 patients~\cite{ge2019predicting}. To improve interpretability, many studies adopt SHAP~\cite{lee2023prediction,ji2023predicting} and LIME~\cite{cho2019predicting}, while others focus on optimizing feature selection~\cite{lai2021exploration} or incorporate rich data sources such as imaging and muscle biosignals~\cite{li2024feature}. However, prior studies mostly rely on feature importance or SHAP-based explanations, often overlooking feature interactions—critical in complex clinical settings. Additionally, many studies are limited by small, single-center datasets. Graph-based models such as Graph Attention Networks have shown promise in addressing challenges associated with EHR data, including heterogeneity and multimodality~\cite{boll2024graph}, and have been successfully applied to other conditions like heart failure and diabetes. Yet, their use in post-stroke outcome prediction remains underexplored.

\paragraph{\textit{Explainability in Clinical Prediction Tasks\\}}

In high-stakes clinical settings, both predictive accuracy and model interpretability are essential. Petch et al.~\cite{petch2022opening} argue that interpretable models should be preferred when they perform comparably to black-box models. Rudin~\cite{rudin2019stop} further criticizes post hoc explanations as unreliable, advocating for inherently interpretable models. However, both inherent and post hoc approaches often struggle to capture complex feature interactions~\cite{carmichael2023well, jiang2023prishap}. For example, acute kidney injury in ICUs—affecting up to 25\% of patients—arises from complex interactions among clinical conditions, not isolated features~\cite{singbartl2012aki}. Some post hoc methods, such as faithful Shapley indices~\cite{tsai2023faith}, and models like NODE-GAM~\cite{lou2013accurate}, address this by incorporating interaction-based explanations. Guo et al.~\cite{guo2024explainable} proposed a graph-based method using attention mechanisms to model feature interactions and generate patient-level explanations. Their method models clinical features as graph components, enabling aggregation from neighbors and capturing dependencies beyond feature importance. Although the interpretability of attention remains debated~\cite{jain2019attention}, we adopt two attention-based GNNs following Guo’s framework, alongside interpretable baselines such as logistic regression and a transformer with SHAP. This design enables comparative analysis of prediction and explanation in post-stroke rigidity modeling.

\section*{Methodology}
Figure~\ref{fig:research_design} shows an overview of the research design. This study employs 519K stroke hospitalization records from the Healthcare Cost and Utilization Project (HCUP) from 2016 to 2020 to develop predictive models for post-stroke rigidity. We utilize a sequential modeling approach by incrementally including additional predictors to understand how prediction accuracy evolves as more clinical information becomes available throughout the hospital journey. For each feature set selection, we implemented five algorithms: Graph-based XAI method with Graphormer~\cite{ying2021transformers}, Graph-based XAI method with Graph Attention Network (GATv2)~\cite{brody2021attentive}, Logistic Regression, XGBoost, and Transformer. We leverage intrinsic explainability metrics for each algorithm (e.g., regression coefficients, tree-based metrics, and attention-based metrics) as well as post-hoc Shapley values to examine changes in feature importance and interactions. Notably, graph-based methods can intrinsically express both feature importance and feature interactions. These approaches enable us to comprehensively understand the predictors of post-stroke rigidity, with graph-based explainability methods (GAT and Graphormer) providing intrinsic feature interaction insights beyond simple feature importance.

\paragraph{\textit{Post-stroke rigidity prediction task.\\}}
For each stroke hospitalization, we generated a binary (0 or 1) target variable \textit{isRigidity} to indicate whether the patient was diagnosed with post-stroke rigidity during hospitalization. Physicians on our team identified 55 specific ICD-10-CM codes (International Classification of Diseases, Tenth Revision, Clinical Modification) indicating post-stroke rigidity, most of which fall under ICD-10-CM G8: Cerebral palsy and other paralytic syndromes. For instance, ICD code G8111, denoting ``Spastic hemiplegia affecting right dominant side," was one of the 55 specific ICD codes used to generate the \textit{isRigidity} label. We found that 43\% of the total 519K stroke hospitalizations had \textit{isRigidity} = 1 (Table~\ref{tab:variables}). After generating the target variable, we removed these 55 specific ICD-10-CM codes from the dataset for prediction purposes. Our task is to use the features listed in Table~\ref{tab:variables}, including information available before/at admission and during hospitalization, to predict the target variable \textit{isRigidity}.

\paragraph{\textit{Data source.\\}}
This study utilizes hospitalization records from the Healthcare Cost and Utilization Project (HCUP-US), comprising 519,000 stroke hospitalizations between 2016 and 2020, where 43\% of cases exhibited rigidity. The study population includes adult patients ($\geq$18 years) with a primary diagnosis of stroke. Table~\ref{tab:variables} outlines the selected features. \textbf{Admission features} encompass demographic factors (age, gender, race), administrative details (payment source, hospital characteristics—division, teaching status, bed size, region, control—transfer status, admission timing—year, month, weekend vs. weekday—admission type—elective vs. non-elective—patient location, and income quartile derived from zip code-based income and education levels), and pre-existing conditions (atrial fibrillation, smoking status, congestive heart failure, diabetes mellitus, hyperlipidemia, hypertension and hypertensive crisis, primary hypertension, chronic kidney disease, previous myocardial infarction, and coronary artery disease). \textbf{In-hospital features} include clinical severity measures (APR-DRG severity score, APR-DRG mortality risk, emergency department utilization, NIH Stroke Scale reporting status, and NIH Stroke Scale score), acute interventions (mechanical thrombectomy and tissue plasminogen activator administration), and ICD diagnosis and procedure codes (top-level ICD codes covering 39 categories). \textbf{Discharge features} consist of transfer-out status, discharge quarter, total charges, uniform disposition code, length of stay, and in-hospital mortality. Among key clinical metrics, the \textbf{NIH Stroke Scale (NIHSS)} is a 15-item neurological examination used to assess acute cerebral infarction effects, evaluating consciousness, language, neglect, visual-field loss, extraocular movement, motor strength, ataxia, dysarthria, and sensory loss. The NIHSS score ranges from 0 to 42, where higher scores indicate greater neurological impairment. The \textbf{APR-DRG severity score} categorizes the extent of physiological decompensation or organ system dysfunction into four levels: minor, moderate, major, and extreme. Similarly, the \textbf{APR-DRG risk of mortality} assigns a subclass to each hospitalization episode, stratified into minor, moderate, major, and extreme, with higher risk of mortality scores correlating with increased mortality risk.\\

\begin{table}[h]
    \centering
    \caption{Selected Variable Descriptions and Distributions}
    \label{tab:variables}
    \scriptsize
    \renewcommand{\arraystretch}{1.3} 
    \begin{tabular}{>{\raggedright\arraybackslash}p{3cm}|>{\raggedright\arraybackslash}p{4.5cm}|>{\raggedright\arraybackslash}p{5.5cm}|c}
        \hline
        \textbf{Variable Category} & \textbf{Variable} & \textbf{Description of Unit of Measure} & \textbf{Distribution, \%} \\
        \hline
        {Target Variables}
        & Rigidity & Binary (1 = patient is diagnosed with rigidity) & 43.45 \\
        \hline
        \multirow{7}{3cm}{\raggedright {Selected variables: Admission Demographics \& Pre-existing Conditions}}
        & AGE & Age in years at admission & 69.85 (14.09) \\
        & HTN & Binary (1 = patient is diagnosed with hypertension) & 82.57 \\
        & DM & Binary (1 = patient is diagnosed with diabetes mellitus) & 32.24 \\
        & HLD & Binary (1 = patient is diagnosed with hyperlipidemia) & 56.59 \\
        & CKD & Binary (1 = patient is diagnosed with chronic kidney disease) & 21.23 \\
        & ELECTIVE & Binary (1 = patient is elective admission) & 4.62 \\
        \hline
        \multirow{2}{3cm}{\raggedright {APRDRG Severity and Risk}}
        & APRDRG\_Severity & All Patient Refined DRG: Severity of Illness Subclass & - \\
        & APRDRG\_Risk\_Mortality & All Patient Refined DRG: Risk of Mortality Subclass & - \\
        \hline
        \multirow{3}{3cm}{\raggedright {Selected variables}: During Hospital Clinical Assessments} 
        & NIHSSREPORT & Binary (1 = patient has an NIHSS assessment recorded) & 39.03 \\
        & NIHSS & NIH Stroke Scale: Objectively quantify the impairment caused by a stroke & 6.96 (7.53) \\
        \hline
        \multirow{4}{3cm}{\raggedright {Selected variables}: During Hospital Procedures/Treatments}
        & MT & Binary (1 = patient received mechanical thrombectomy treatment) & 3.91 \\
        & TPA & Binary (1 = patient received tissue plasminogen activator treatment) & 8.09 \\
        \hline
        \multirow{15}{3cm}{\raggedright {Selected variables}: During Hospital Top-Level ICD Diagnosis \& Procedure Codes} 
        & has\_diag\_Symptoms\_and\_signs & Binary (1 = patient has a diagnosis with ICD codes starting with R00–R99) & 78.39 \\
        & has\_diag\_Digestive\_system\_diseases & Binary (1 = patient has a diagnosis with ICD codes starting with K00–K95) & 29.61 \\
        & has\_diag\_Eye\_diseases & Binary (1 = patient has a diagnosis with ICD codes starting with H00–H59) & 16.06 \\
        & has\_diag\_Nervous\_system\_diseases & Binary (1 = patient has a diagnosis with ICD codes starting with G00–G99) & 46.65 \\
        & has\_diag\_Infectious\_and\_parasitic\_diseases & Binary (1 = patient has a diagnosis with ICD codes starting with A00–B99) & 15.04 \\
        & has\_pr\_Medical\_and\_Surgical & Binary (1 = patient has a procedure related to medical and surgical) & 30.19 \\
        & has\_pr\_Extracorporeal\_or\_Systemic\_Assistance & Binary (1 = patient has a procedure related to extracorporeal or systemic assistance and performance) & 13.42 \\
        & has\_pr\_Obstetrics & Binary (1 = patient has a procedure related to obstetrics) & 0.019 \\
        \hline
    \end{tabular}
\end{table}

\paragraph{\textit{Incremental modeling strategy.\\}} 
We implemented an incremental modeling strategy that mirrors the temporal availability of features in clinical settings (Selected features are in Table \ref{tab:variables}). This approach allows for prediction at different time points in the patient care continuum while addressing the inherent challenge of temporal relationships in diagnosis codes. This incremental modeling strategy explicitly addresses the trade-off between model complexity and clinical utility. By sequentially incorporating variables according to their temporal availability in the hospitalization journey, we can determine: (1) The earliest point at which meaningful prediction becomes possible, and (2) better analyze the contribution of each factor in the prediction. The specific models are as follows: \texttt{Model 1} uses Admission Information and pre-existing conditions as predictors. Building upon Model 1, \texttt{Model 2} further adds In-Hospital Assessment and Treatment. In \texttt{Model 3}, we added In-hospital ICD diagnosis codes and procedure codes. This model integrates all ICD diagnosis and procedure codes with 39 top-level ICD codes. \texttt{Model 4} further incorporates discharge-related variables. Note that discharge information is collected after the rigidity diagnosis. Table~\ref{tab:feature_importance} shows the results of each model.



\paragraph{\textit{Predictive algorithms and explainability mechanism selections.\\}}
We implemented five machine learning algorithms across all feature configurations and applied appropriate explainability mechanisms (Figure \ref{fig:research_design}). The dataset was split with 80\% (415K hospitalizations) for training and validation and 20\% (103K hospitalizations) for testing. We used validation to monitor performance and adjust hyperparameters. Models were evaluated using standard classification metrics including precision, recall, F1 score, ROC-AUC, and area under the precision-recall curve (AUPRC). We also reported sensitivity and specificity given their importance in clinical decision-making.

\paragraph{\texttt{1. Graph-based Explainable AI Method.}} 
We employed a graph-based explainable AI framework introduced by Guo~\cite{guo2024explainable}, which is \textit{\textbf{intrinsically interpretable}}. This framework captures feature interactions through an internal attention mechanism rather than relying on post-hoc approximations. For each hospitalization, we constructed a fully connected, directed graph where nodes represent clinical features and edges encode information flow between features. This patient-specific approach enables customized explanations while improving prediction accuracy. We implemented this framework using two graph-transformer-based methods: \texttt{Graphormer} and \texttt{Graph Attention Networks} (GAT). For each patient admission, we: (1) Constructed a fully connected directed graph with nodes representing input features; (2) Transformed each node into a 64-dimensional vector using a feedforward neural network; (3) Calculated edge importance using attention mechanisms—GATv2 computes attention weights using:

\[
\alpha_{ji} = \text{softmax}\left(\mathbf{a}^{\mathsf{T}} \operatorname{LeakyReLU} \left(\mathbf{W}_{\text{left}}\mathbf{h}_i + \mathbf{W}_{\text{right}}\mathbf{h}_j\right) \right)
\]

while Graphormer uses dot product attention. We then (4) Calculated node importance using global attention pooling to aggregate node embeddings into a graph-level representation, with each node assigned an attention weight \(\beta_i\); (5) Generated rigidity predictions using an MLP with softmax; and (6) Extracted explanations as feature importance (\(\text{featimp}_i = \beta_i\)) and feature interaction importance (\(\text{Intimp}_{ji} = \alpha_{ji} \times \text{featimp}_i\)).

\paragraph{\texttt{2. Logistic Regression.}} 
We implemented Logistic Regression with L2 regularization, addressing class imbalance through class weighting. The model used the 'liblinear' solver with a maximum of 1000 iterations. For interpretability, we computed SHAP (SHapley Additive Explanations) values to quantify each feature's contribution.

\paragraph{\texttt{3. XGBoost.}} 
We employed XGBoost to capture complex nonlinear relationships, using scale-pos-weight to address class imbalance. The model was configured with a learning rate of 0.001, maximum tree depth of 10, and 100 boosting iterations, with logarithmic loss monitored for hyperparameter tuning. For interpretability, we extracted feature importance based on split frequency across trees and computed global SHAP values to quantify each feature's overall contribution to predictions.

\paragraph{\texttt{4. Transformer.}} 
Our transformer architecture included a linear embedding layer projecting input features to 128 dimensions, followed by a 5-layer transformer encoder with 8 attention heads per layer. Each encoder incorporated multi-head self-attention with position-wise feed-forward networks (ReLU activation, 4× hidden dimension multiplier) and dropout regularization (0.3). The classification head consisted of a two-layer MLP that reduced dimensions from 128 to 64 with ReLU activation, ending with a sigmoid unit for binary prediction. We trained using BCE loss with positive class weighting for 50 epochs (batch size 256), employing a warmup learning rate scheduler (initial rate 3e-4) and adaptive threshold optimization. For explainability, we implemented permutation-based feature importance—measuring performance decrease after feature shuffling—and SHAP values calculated using KernelExplainer. The permutation method captured magnitude effects independently, while SHAP values provided both magnitude and directional insights while accounting for feature interactions.

\section*{Results}
\paragraph{\textit{Predictive performance comparison across models and algorithms.\\}}
As shown in Table~\ref{tab:performance}, we have two key observations. First, the benefits of the gradual modeling strategy indicate that during-hospital clinical assessments are the best predictors for rigidity prediction. From Model 1 (with only information at the admission stage) to Model 4 (with admission information, during-hospital clinical assessments, ICD codes, and discharge information), the prediction performance incrementally increased. We witnessed the largest boost (approximately 11\% across all models) from Model 2, when adding the during-hospital clinical assessments, specifically: \textit{(1) Clinical severity measures}: APR-DRG severity score, APR-DRG mortality risk, emergency department utilization, NIH Stroke Scale reporting status, and NIH Stroke Scale score (NIHSS). \textit{(2) Acute interventions}: Mechanical thrombectomy and tissue plasminogen activator administration. From Model 2 to Model 3, the average accuracy increased by approximately 1\%, and from Model 3 to Model 4, the average increase was around 1\%. During-hospital assessments are the key contributors to rigidity prediction.\\

\begin{table}[h]
    \centering
    \caption{Predictive Performance Comparison}
    \label{tab:performance}
    \scriptsize
    \renewcommand{\arraystretch}{1.3} 
    \begin{tabular}{p{1cm}|p{4cm}|p{2.1cm} c c c c c c}
        \hline
        \textbf{Models} & \textbf{Sequential Feature Sets} & \textbf{Algorithms} & \textbf{Accuracy} & \textbf{AUROC} & \textbf{AUPRC} & \textbf{F1} & \textbf{Specificity} & \textbf{Sensitivity} \\
        \hline
        \multirow{5}{1cm}{\scriptsize Model 1}
        &
        \multirow{5}{4cm}{\raggedright  \scriptsize
        \textbf{\textit{(1) Admission:}} Admin Info \& Pre-existing Conditions}
        & Logistic Regression & 0.547 & 0.566 & \textbf{0.492} & \textbf{0.514} & 0.546 & \textbf{0.551} \\
        & & XGBoost & 0.541	& 0.555	& 0.481	& 0.497	& 0.556	& 0.522 \\
        & & Transformer & 0.554	& \textbf{0.567}	& \textbf{0.492}	& 0.500	& 0.585 & 0.513 \\
        & & Graphormer & \textbf{0.572}	& 0.560 & 0.485	& 0.236 &\textbf{0.894} & 0.152 \\
        & & GATv2 & 0.567	& 0.555	& 0.481	& 0.343	& 0.802	& 0.260 \\
        \hline
        \multirow{5}{1cm}{\scriptsize Model 2}
        &
        \multirow{5}{4cm}{\raggedright \scriptsize
        \textbf{\textit{(1) Admission:}} Admin Info \& Pre-existing Conditions \\  [0.3mm]
        + \textbf{\textit{(2) Hospital:}} Clinical Assessments}
        & Logistic Regression & 0.654 & 0.692 &  0.639 & 0.582	& 0.730	& 0.555 \\
         & & XGBoost & 0.663	& 0.712	& 0.659	& 0.602	& 0.721	& 0.587 \\
         & & Transformer & 0.663	& 0.714	& 0.657	& \textbf{0.612}	& 0.702	& \textbf{0.613} \\
         & & Graphormer & 0.669	& 0.714	& 0.661	& 0.581	& 0.775	& 0.530 \\
         & & GATv2 & \textbf{0.673}	& \textbf{0.719}	& \textbf{0.667}	& 0.586	& \textbf{0.779}	& 0.534 \\
        \hline
        \multirow{5}{1cm}{ \scriptsize Model 3}
        &
        \multirow{5}{4cm}{\raggedright\scriptsize
        \textbf{\textit{(1) Admission:}} Admin Info \& Pre-existing Conditions \\[0.3mm]
        \textbf{\textit{(2) Hospital:}} Clinical Assessments \\[0.3mm]
        + \textbf{\textit{(3) Hospital:}} Top-Level ICD Diagnosis \& Procedure Codes
        }
        & Logistic Regression & 0.661 & 0.704	& 0.648	& 0.596	& 0.725	& 0.577 \\
         & & XGBoost & 0.666	& 0.725	& 0.672	& 0.619	& 0.698	& 0.625 \\
         & & Transformer & 0.672	& 0.731	& 0.674	& \textbf{0.625}	& 0.619	& \textbf{0.631} \\
         & & Graphormer & 0.677	& 0.727	& 0.671	& 0.590	& \textbf{0.785}	& 0.536 \\
         & & GATv2 & \textbf{0.681}	& \textbf{0.736}	& \textbf{0.681}	& 0.611	& 0.761	& 0.577 \\
        \hline
        \multirow{5}{1cm}{ \scriptsize Model 4}
        &
        \multirow{5}{4cm}{\raggedright \scriptsize
        \textbf{\textit{(1) Admission:}} Admin Info \& Pre-existing Conditions \\[0.1mm]
        \textbf{\textit{(2) Hospital:}} Clinical Assessments \\[0.1mm]
         \textbf{\textit{(3) Hospital:}} Top-Level ICD Diagnosis \& Procedure Codes\\[0.1mm]
        +  \textbf{\textit{(4) Discharge:}} Admin Outcomes}
        & Logistic Regression &0.670 &0.724 &0.668 &0.617 &0.714 &0.612 \\
         & & XGBoost &0.675 &0.738 &0.682 &0.638 &0.686 &0.661 \\
         & & Transformer &0.684 &0.746 &0.688 &0.640 &0.713 &0.646 \\
         & & Graphormer &0.683 &0.743 &0.687 &0.621 &\textbf{0.747} &0.599 \\
         & & GATv2 &\textbf{0.687} &\textbf{0.752} &\textbf{0.699} &\textbf{0.648} &0.705 &\textbf{0.664} \\
        \hline
    \end{tabular}
\end{table}
Second, for each model, the performance across all the algorithms (from logistic regression to graph-based models) is similar. When the performance of algorithms is similar, models with better explainability are more desirable. One thing to note is that discharge information is not helpful for predicting in-hospital post-stroke outcomes. Therefore, we will mainly focus on Model 3: admission information + pre-existing conditions + in-hospital clinical assessments + top-level ICD codes for investigating model explainability. Through the analysis below, we can understand how specific clinical assessments predict the rigidity outcome and how the predicting features influence each other. At the same time, we need to keep in mind that the better the predictive performance, the better the modeling, and potentially better explainable insights. As the rigidity prediction is a complex task, the results in this section will be more on the correlation side and may include randomness and need to be evaluated with domain knowledge.

\paragraph{\textit{Feature importance \& feature interactions.\\}}
Feature importance in Table~\ref{tab:feature_importance} further confirms how certain factors during hospital clinical assessments contribute to rigidity prediction. We chose the feature settings of Model 3, i.e., \textit{(1) Admission:} Admin Info \& Pre-existing Conditions, \textit{(2) Hospital:} Clinical Assessments, \textit{(3) Hospital:} Top-Level ICD Diagnosis \& Procedure Codes. Across all the models, we report the top ten features with the highest feature importance. For logistic regression, XGBoost, and transformer models, we report SHAP values, while graph-based models incorporate node-level attention and averaged outward edge attention (i.e., the contribution of one feature to others). Graph-based models offer an advantage: rather than assessing feature importance in isolation, they capture how features influence each other. Among the top ten most influential features across at least three of the five models, three during hospital assessments stand out as the strongest predictors—APRDRG Severity, APRDRG Risk Mortality, and NIH Stroke Scale. Analysis of both feature importance and SHAP values further supports this ranking, revealing that these three features consistently appear as highly influential across both metrics. For instance, XGBoost feature importance scores for these features were 0.204, 0.165, and 0.140, respectively, while their corresponding mean absolute SHAP values were 0.020, 0.036, and 0.021. Furthermore, we computed the Spearman correlation coefficient between XGBoost feature importance and SHAP values, which resulted in $\rho = 0.806$ ($p = 0.0049$), indicating strong alignment between these metrics. This stability underscores the clinical relevance of these predictors, particularly in identifying high-risk patients.\\

\begin{table}[h]
    \centering
    \caption{Salient features identified by Model 3. The top 10 contributing features are listed. The bold features are those that appear in at least four out of five models.}
    \label{tab:feature_importance}
    \scriptsize
    \renewcommand{\arraystretch}{1.5}
    \begin{tabularx}{\textwidth}{X|>{\centering\arraybackslash}m{1.5cm}|X|>{\centering\arraybackslash}m{1.5cm}|X|>{\centering\arraybackslash}m{1.5cm}}
        \hline
        \multicolumn{2}{c|}{\textbf{Logistic Regression}} & \multicolumn{2}{c|}{\textbf{XGBoost}} & \multicolumn{2}{c}{\textbf{Transformer}}\\
        \cline{2-6}
        \hline
        Feature Name & Shapley Value & Feature Name & Shapley Value & Feature Name & Shapley Value\\
        \hline
        \textbf{APRDRG Severity} & 0.5490 & \textbf{NIHSS} & 0.0362 & NIHSSREPORT & 0.0596\\
        \textbf{APRDRG\_Risk Mortality} & 0.5336 & \textbf{APRDRG\_Risk Mortality} & 0.0209 & \textbf{APRDRG\_Risk Mortality} & 0.0503\\
        NIHSSREPORT & 0.2796 & \textbf{APRDRG Severity} & 0.0200 & \textbf{APRDRG\_Severity} & 0.0491\\
        \textbf{NIHSS} & 0.1848 & \textbf{has\_diag\_Symptoms and signs} & 0.0097 & \textbf{NIHSS} & 0.0400\\
        \textbf{has\_diag\_Symptoms and signs} & 0.1417 & TPA & 0.0069 & \textbf{has\_diag\_Symptoms and signs} & 0.0110\\
        TPA & 0.1123 & \textbf{has\_diag\_Eye diseases} & 0.0065 & TPA & 0.0103\\
        \textbf{has\_diag\_Nervous system diseases} & 0.0935 & MT & 0.0055 & MT & 0.0080\\
        \textbf{has\_diag\_Eye diseases} & 0.0932 & \textbf{has\_diag\_Nervous system \newline diseases} & 0.0031 & HOSP\_LOCTEACH\_2 & 0.0062\\
        MT & 0.0796 & has\_diag\_Infectious and \newline parasitic diseases & 0.0028 & \textbf{has\_diag\_Eye diseases} & 0.0061\\
        RACE & 0.0658 & AGE & 0.0021 & \textbf{has\_diag\_Nervous system \newline diseases} & 0.0045\\
        \hline
    \end{tabularx}

        \begin{tabularx}{\textwidth}{>{\raggedright\arraybackslash}m{2.32cm}|>{\centering\arraybackslash}m{0.93cm}|>{\raggedright\arraybackslash}m{2.32cm}|>{\centering\arraybackslash}m{0.93cm}|>{\raggedright\arraybackslash}m{2.32cm}|>{\centering\arraybackslash}m{0.93cm}|>{\raggedright\arraybackslash}m{2.32cm}|>{\centering\arraybackslash}m{0.93cm}}

        \hline
        \multicolumn{4}{c|}{\textbf{Graphormer}} & \multicolumn{4}{c}{\textbf{GATv2}} \\
        \cline{2-8}
        \hline
        Feature Name & Average Node Attention & Feature Name & Average Outward Edge Attention & Feature Name & Average Node Attention & Feature Name & Average Outward Edge Attention\\
        \hline
        \textbf{APRDRG\_Risk Mortality} & 0.1915 & ELECTIVE & 0.1533 & \textbf{APRDRG\_Risk Mortality} & 0.1479 & \textbf{APRDRG\_Severity} & 0.2618 \\
        \textbf{APRDRG\_Severity} & 0.1004 & \textbf{has\_diag\_Eye diseases} & 0.1461 & YEAR & 0.0836 & \textbf{has\_diag\_Symptoms and signs} & 0.1025 \\
        \textbf{NIHSS} & 0.0984 & PL\_NCHS & 0.0983 & NIHSSREPORT & 0.0617 & \textbf{NIHSS} & 0.0728 \\
        \textbf{has\_diag\_Symptoms and signs} & 0.0848 & has\_diag\_Infectious and parasitic diseases & 0.0898 & has\_diag\_Infectious and parasitic diseases & 0.0496 & has\_diag\_Digestive system diseases & 0.0522 \\
        PAY1 & 0.0580 & \textbf{has\_diag\_Symptoms and signs} & 0.0605 & has\_diag\_Injuries and poisonings & 0.0494 & NIHSSREPORT & 0.0482 \\
        \textbf{has\_diag\_Nervous system diseases} & 0.0397 & has\_diag\_Circulatory system diseases & 0.0515 & \textbf{has\_diag\_Eye diseases} & 0.0407 & PAY1 & 0.0307 \\
        PL\_NCHS & 0.0356 & RACE & 0.0427 & HOSP\_DIVISION & 0.0393 & RACE & 0.0290 \\
        HOSP\_DIVISION & 0.0293 & has\_diag\_Musculoskeletal diseases & 0.0391 & has\_diag\_Genitourinary diseases & 0.0349 & HTN & 0.0271 \\
        has\_diag\_Genitourinary diseases & 0.0224 & has\_diag\_Congenital malformations & 0.0329 & has\_pr\_Medical and Surgical & 0.0330 & \textbf{has\_diag\_Nervous system diseases} & 0.0245 \\
        DM & 0.0221 & has\_diag\_Ear diseases & 0.0309 & AGE & 0.0326 & has\_diag\_Neoplasms & 0.0227\\
        \hline
    \end{tabularx}
\end{table}

Graph-based models reveal additional insights through feature interaction pathways. The overall feature importance (node attention and the top outwards attention weights) of graph-based algorithms aligns with traditional machine learning models (logistic regression, XGBoost, and transformer), but we can identify some potential feature interaction pathways. From Table~\ref{tab:feature_importance} and Table~\ref{feature interaction} graphormer results, we can tell some administrative information contributes to the critical clinical assessment, e.g., Elective (if the admission was scheduled) contributes to the APRDRG Risk Mortality, which is reasonable as unscheduled admission indicates a more severe situation. Furthermore, PL\_NCHS (a classification of Patient Location, indicating the Urban-Rural classification), influences the APRDRG Risk Mortality. Also, the ``has diag Symptoms and signs" (if the patient has diagnosis codes of symptoms and signs, which are ICD codes starting with R00–R99), also contributes to the APRDRG Risk Mortality. Additionally, our GATv2 implementation finds some similar feature interactions (e.g.,``has diag Symptoms and signs" interacts with APRDRG Risk Mortality), but it also finds some different pathways, for example, the interaction between NIH Stroke score and APRDRG Severity, which makes sense. These different nuances may have two reasons: first is the complex nature of post-stroke rigidity—there are multiple pathways in nature; second may be due to our current data limitations (e.g., a limited number of clinical assessment information) which prevents some dominant feature interactions from being found. But we believe this intrinsically interpretable ability is a good property. For instance, instead of only understanding what the key contributors of rigidity are, we can also see the interactions between the features, getting a better understanding for clinical decision making.\\

\begin{table}[h]
    \centering
    \caption{Salient feature interactions. The top 10 contributing feature interactions are listed.}
    \label{feature interaction}
    \renewcommand{\arraystretch}{1.3}
    \setlength{\tabcolsep}{4pt}
    \scriptsize
    \begin{tabularx}{\textwidth}{X|X|c|X|X|c}
        \hline
        \multicolumn{3}{c|}{\textbf{Graphormer}} & \multicolumn{3}{c}{\textbf{GATv2}} \\
        \hline
        \textbf{Feature Name \newline(Source)} & \textbf{Feature Name \newline(Destination)} & \textbf{Edge Attention} & 
        \textbf{Feature Name \newline(Source)} & \textbf{Feature Name \newline(Destination)} & \textbf{Edge \newline Attention} \\
        \hline
        PL\_NCHS & APRDRG\_Risk\_Mortality & 0.0051 & APRDRG\_Severity & APRDRG\_Risk\_Mortality & 0.0629 \\
        ELECTIVE & APRDRG\_Risk\_Mortality & 0.0049 & \makecell[l]{diag\_Symptoms and \\signs} & APRDRG\_Risk\_Mortality & 0.0242 \\
        \makecell[l]{has\_diag\_Symptoms and \\signs} & APRDRG\_Risk\_Mortality & 0.0048 & APRDRG\_Severity & YEAR & 0.0188 \\
        has\_diag\_Eye diseases & NIHSS & 0.0047 & diag\_Digestive system diseases & APRDRG\_Risk\_Mortality & 0.0170 \\
        has\_diag\_Eye diseases & \makecell[l]{has\_diag\_Symptoms and \\signs} & 0.0047 & APRDRG\_Severity & NIHSSREPORT & 0.0148 \\
        has\_diag\_Eye diseases & APRDRG\_Severity & 0.0045 & APRDRG\_Severity & diag\_Eye diseases & 0.0142 \\
        ELECTIVE & APRDRG\_Severity & 0.0044 & APRDRG\_Risk\_Mortality & YEAR & 0.0141 \\
        ELECTIVE & \makecell[l]{has\_diag\_Symptoms and \\signs} & 0.0043 & APRDRG\_Severity & diag\_Infectious and parasitic diseases & 0.0119 \\
        has\_diag\_Musculoskeletal diseases & APRDRG\_Risk\_Mortality & 0.0043 & APRDRG\_Severity & diag\_Injuries and poisonings & 0.0117 \\
        PL\_NCHS & APRDRG\_Severity & 0.0043 & NIHSS & NIHSSREPORT & 0.0090 \\
        \hline
    \end{tabularx}
\end{table}
\begin{wrapfigure}{r}{0.6\textwidth} 
    \centering
    \vspace{-10pt}  
    \includegraphics[width=\linewidth]{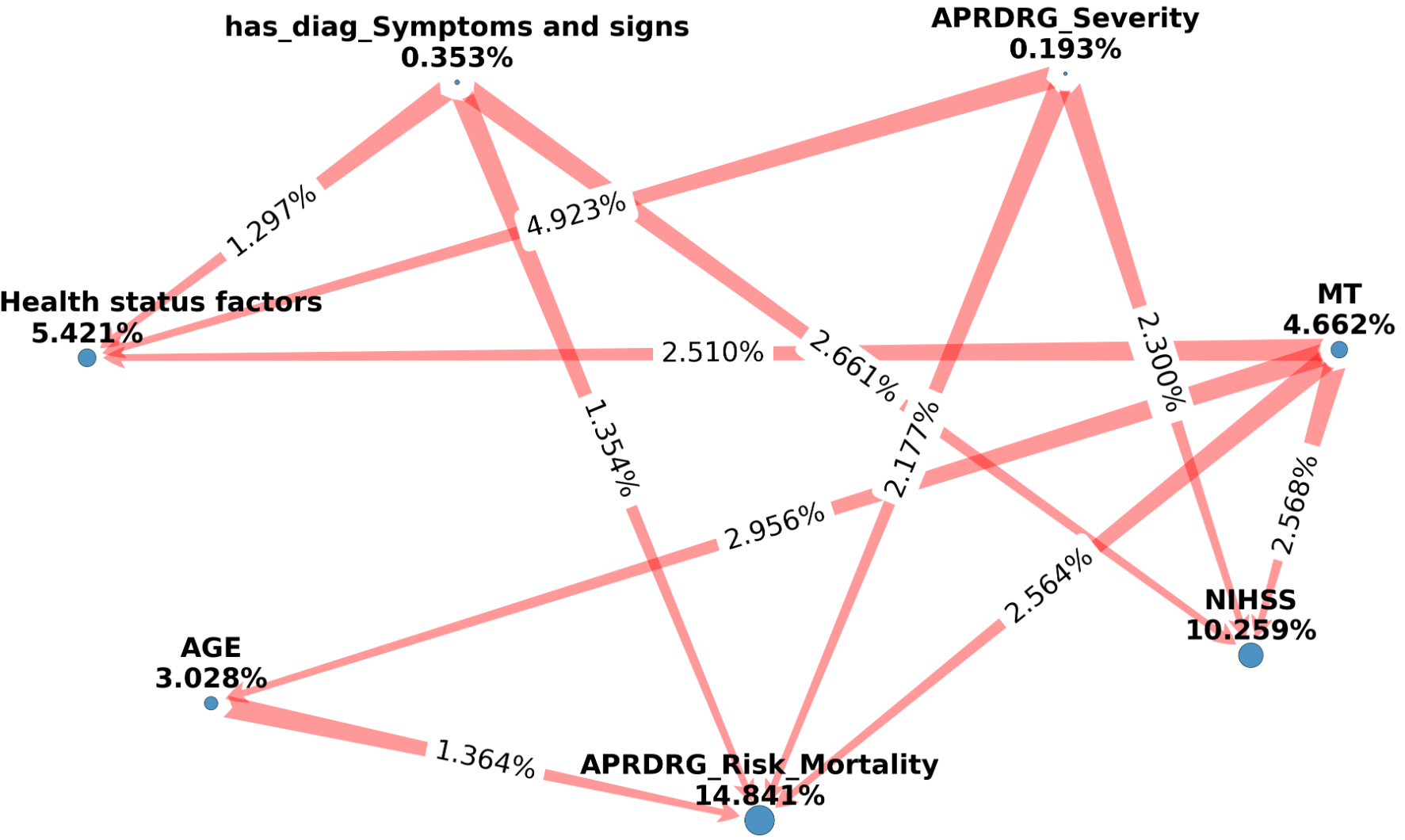}
    \captionsetup{skip=0pt}
    \caption{Personalized explanations from the Graph Learning Model using GATv2.}
    \label{fig:Personalized_Explanations_Graphormer}
\end{wrapfigure}

\paragraph{\textit{Individual-level explanations.\\}}

Graph-based models also exhibit significant potential for individual-level customized explanations. For each post-stroke hospitalization, we can derive the corresponding likelihood of post-stroke rigidity, feature importance (node attention), and feature interactions. Figure~\ref{fig:Personalized_Explanations_Graphormer} presents an example of a male patient with Mechanical Thrombectomy (MT) treatment, NIH Stroke Scale = 19 (moderate to severe stroke), APR-DRG Risk Mortality classified as group 2 (moderate), APR-DRG Severity also classified as group 2 (moderate), and post-stroke rigidity (with the graph model predicting an 87.4\% likelihood of rigidity). With 73 predictive features, there are a total of 73×73=5,329 potential feature interactions. For clarity, we filtered out 99.8\% of the links, retaining only the 11 connections (0.2\%) with the highest feature interaction values, calculated as the product of normalized edge attention weights and destination feature weights. The node size and label represent each node’s contribution to the final rigidity prediction, ensuring that the sum of all node attention values equals 1. In this specific case (Figure~\ref{fig:Personalized_Explanations_Graphormer}), the NIH Stroke Scale contributes 10.259\% to the final prediction, while APR-DRG Risk Mortality contributes 14.841\%. Each edge represents the proportion that the source node contributes to the final representation of the destination node, with all 5,329 edge importance values summing to 1. For example, the edge from MT to the NIH Stroke Scale, labeled 2.568\%, indicates that 25\% of NIHSS’s total contribution (i.e., 2.568\%/10.259\%) originates from MT. This analysis enables us to trace how MT, APR-DRG Severity, and AGE contribute to key predictive features, such as the NIH Stroke Scale and APR-DRG Risk Mortality. These individualized explanations provide clinicians with a deeper understanding of a patient’s risk profile, facilitating more personalized and targeted interventions.

\paragraph{\textit{Qualitative feedback from a stroke specialist\\}}
We invited a stroke specialist to provide qualitative feedback on our predictive approach for post-stroke rigidity, specifically regarding feature importance and feature interactions. Compared to baseline explanation methods, which only report feature importance, the graph-based method demonstrates promising potential in capturing feature interactions when evaluating individual hospitalizations. For instance, all five models consistently identify the NIH Stroke Scale, APR-DRG Risk Mortality, and APR-DRG Severity scores as the most predictive features for post-stroke rigidity. However, as illustrated in Figure~\ref{fig:Personalized_Explanations_Graphormer}, the graph-based method further reveals how MT and age contribute to these key features—specifically, the NIH Stroke Scale and APR-DRG Risk Mortality. This insight is clinically meaningful, as age is a well-established risk factor for stroke outcomes, and MT may influence neurological recovery and post-stroke complications through vascular or inflammatory pathways. By extending beyond traditional feature importance analysis, the graph-based XAI method provides deeper insights into the complex interplay between predictive factors for post-stroke rigidity. This enhanced interpretability holds significant potential for guiding personalized interventions. Future research should incorporate more extensive longitudinal data and undergo broader expert evaluations to further validate the clinical utility of this approach.

\subsection*{Conclusion}
This study investigates the consistency between graph-based explainable AI models and traditional machine learning explainability techniques such as SHAP. Our findings consistently highlighted the central role of during-hospital clinical assessments, particularly the NIH Stroke Scale, APR-DRG risk of mortality, and APR-DRG severity score, in predicting post-stroke rigidity. Notably, graph-based models provide a unique advantage: they not only confirm key predictors identified by traditional models but also reveal meaningful feature interactions at both the population and individual levels, enhancing interpretability for clinical decision-making.

Despite these promising results, our study has several limitations. The best model performance achieves only 68\% accuracy and 0.75 AUROC, reflecting the inherent complexity of rigidity prediction. Additionally, richer feature sets—incorporating more detailed clinical assessments—could further improve both predictive accuracy and the granularity of feature interaction insights. Future work should explore additional data sources and validation across broader patient populations to strengthen the robustness and applicability of these findings. The debate on whether attention metrics can be interpreted as explainability is ongoing~\cite{jain2019attention}; more effort is needed to develop reliable XAI methods that excel in both interpretability and performance.

\subparagraph{Acknowledgments}
We would like to acknowledge the following funding supports: NIH OT2OD032581, NIH OTA-21-008, NIH 1OT2OD032742-01.

\makeatletter
\renewcommand{\@biblabel}[1]{\hfill #1.}
\makeatother

\bibliographystyle{vancouver}
\bibliography{manuscript}





\end{document}